\documentclass[letterpaper, 10 pt, conference]{ieeeconf}  

\IEEEoverridecommandlockouts                              

\usepackage{cite}
\usepackage{amsmath,amssymb,amsfonts}
\usepackage{algorithmic}
\usepackage{graphicx}
\usepackage{textcomp}
\usepackage{xcolor}

\usepackage{graphics} 
\usepackage{epsfig} 
\usepackage{mathptmx} 
\usepackage{times} 
\usepackage{amsmath} 
\usepackage{bm}
\usepackage{amssymb}  
\usepackage{newtxtext, newtxmath}
\usepackage{hyperref}

\usepackage{algorithmic}
\usepackage{booktabs}
\usepackage{adjustbox}
\usepackage{multirow}
\usepackage[lined,boxed,ruled,noend]{algorithm2e}
\usepackage{balance}

\usepackage{xcolor}

\newcommand{\abbv}{\texttt{EMOTION}\xspace}
\newcommand{\abbvh}{\texttt{EMOTION++}\xspace}
\newcommand{\para}[1]{\noindent\textbf{#1.}}

\title{\LARGE \bf
\abbv: Expressive Motion Sequence Generation for Humanoid Robots with In-Context Learning
}

\author{Peide Huang$^{1}$ Yuhan Hu$^{1}$ Nataliya Nechyporenko$^{1}$ Daehwa Kim$^{1}$ Walter Talbott$^{1}$ Jian Zhang$^{1}$
\thanks{$^{1}$PH, YH, NN, DK, WT, JZ are with Apple, California, USA. Corresponding to Peide Huang: 
        {\tt\small peide\_huang@apple.com}}%
}

\begin{document}

\maketitle
\thispagestyle{empty}
\pagestyle{empty}

\begin{abstract}

This paper introduces a framework, called \abbv, for generating expressive motion sequences in humanoid robots, enhancing their ability to engage in human-like non-verbal communication. Non-verbal cues such as facial expressions, gestures, and body movements play a crucial role in effective interpersonal interactions. Despite the advancements in robotic behaviors, existing methods often fall short in mimicking the diversity and subtlety of human non-verbal communication. To address this gap, our approach leverages the in-context learning capability of large language models (LLMs) to dynamically generate socially appropriate gesture motion sequences for human-robot interaction. We use this framework to generate 10 different expressive gestures and conduct online user studies comparing the naturalness and understandability of the  motions generated by \abbv and its human-feedback version, \abbvh, against those by human operators. The results demonstrate that our approach either matches or surpasses human performance in generating understandable and natural robot motions under certain scenarios. We also provide design implications for future research to consider a set of variables when generating expressive robotic gestures. 
\end{abstract}


\section{INTRODUCTION}

People use non-verbal communication to interact with others all the time. Facial expressions, gestures, and body movements are key components of effective interpersonal communication \cite{Urakami2022-no, salem201723}. These unspoken elements can convey emotions and intentions more profoundly than verbal means alone. As such, integrating non-verbal communication capabilities into robots can potentially lead to more natural and pleasant human-robot interactions.

Previous research has highlighted the importance of expressive behaviors in robots and investigated different methods for creating these behaviors across a range of applications and settings. Studies have shown that robots capable of displaying human-like emotions and reactions can improve user satisfaction and engagement\cite{saunderson2019robots, Zabala2021-zz}. Techniques ranging from pre-defined sequences to dynamic, sensor-driven behaviors have been implemented \cite{Takayama2011-ww}, yet they often lack the adaptability required for real-world interactions that mimic human diversity and subtlety.

As humanoid is the frontier embodiment of general-purpose robots, it is critical to enable humanoid robots to generate expressive gesture motions. These motions need to be contextually appropriate and vary dynamically with the social environment and interaction goals. Given the unlimited number of diverse and subtle gestures, existing work that relies on human-engineered motion primitives or pre-recorded trajectories demands tremendous efforts and resources~\cite{Gross2017-zj, gray2010expressive}. Therefore, it is essential to develop a method where the robots can interpret human social cues and respond with natural and engaging gestures with minimal human efforts.\looseness=-1

\begin{figure*}[t]
\centering
\includegraphics[width=0.99\textwidth]{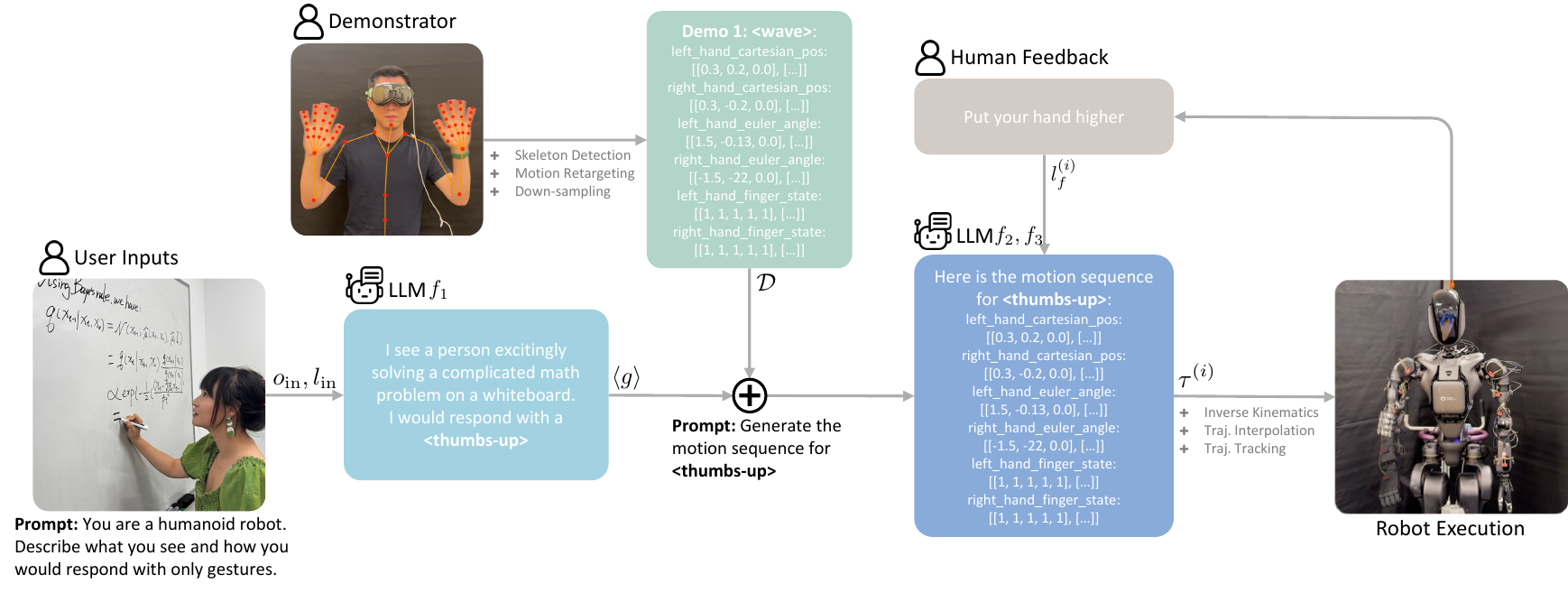}
 \vspace{-0.3cm}
\caption{Overview of the \abbv framework.}
 \vspace{-0.4cm}
\label{fig:overview}
\end{figure*}

Leveraging the in-context learning ability of large language models (LLMs) and vision-language models (VLMs)~\cite{brown2020language}, we propose a new approach, \abbv, to generate expressive motion sequences for humanoid robots under diverse social contexts. Our method utilizes the contextual understanding and sequence generation capabilities of these models to produce a set of expressive motion sequences. These sequences are not only varied and rich but also tailor-made for specific interaction scenarios, enabling robots to behave in a way that feels natural and understandable to users.\looseness=-1

In an online user study, we demonstrate that \abbv and \abbvh are comparable to or even better than the human oracle in terms of naturalness and understandability. Participants were asked to evaluate videos of robot motion generated by our methods versus those demonstrated by a human performing the same gestures. 
We also found that \abbvh are rated as significantly better than \abbv regarding both naturalness and understandability, suggesting that integrating human feedback could improve the quality of the generated gestures.
Additionally, we found that there is a high variance of the perception across different gestures, while some gestures received high ratings ($>5$), such as ``stop'', and ``thumbs-up'', a few gestures were less well-received ($<3$) such as ``listening'', and ``jazz-hands''.
Furthermore, the qualitative reasoning offers insights into which robotic variables influence human perception of expressive gestures. We highlight key factors such as hand position, movement patterns, arm and shoulder articulation, finger pose, and speed.\looseness=-1

Overall, the results support our approach, emphasizing the effectiveness of using LLMs and VLMs for enhancing humanoid robot expressiveness.
We also provide design implications for future research to consider a set of variables when generating expressive robotic gestures.

\section{RELATED WORK}
\para{Expressive Human-Robot Interactions}
Human communication encompasses a broad range of non-verbal cues. Translating these cues into robot design can facilitate more natural, engaging, and accessible interactions between robots and humans \cite{Urakami2022-no, salem201723}. Previous research has shown that robots can effectively utilize non-verbal cues, such as body language \cite{Zabala2021-zz, salem2011friendly}, proxemics \cite{rios2015proxemics}, facial expressions, and eye gaze \cite{admoni2017social}, to enhance human-robot interactions. These expressive behaviors can improve communication, foster empathy, increase user interest, and enhance the perceived intelligence and friendliness of robots \cite{saunderson2019robots, Zabala2021-zz, bethel2007survey}.

Traditionally, robot designers and animators have collaborated to manually craft expressive and dynamic robot behaviors \cite{Gross2017-zj, gray2010expressive}. Prior works use heuristic approaches to map human motions from video footage, demonstrations, or animated characters, translating these behaviors to the robot’s embodiment \cite{Takayama2011-ww}. However, this process demands tremendous efforts and requires a high level of expertise in different embodiments.\looseness=-1

Recently, generative models have been utilized to create or modify expressive robot behaviors from smaller sets of demonstrated samples \cite{Suguitan2020-we, 10.1016/j.robot.2018.11.024}. These models can significantly reduce the effort required from human demonstrators and have the potential to produce a more diverse range of expressive behaviors, leading to greater efficiency and adaptability.



Recent advancements in LLMs have significantly improved the generation of natural and expressive behaviors in human-robot interactions \cite{Wei2024-ab, liang2023code}. Traditional rule-based and template-based methods, which struggle to produce complex, multimodal behaviors and require extensive human intervention or specialized datasets. In contrast, LLMs allow for more adaptive and diverse behavior generation with minimal input \cite{zhang2023large}, as demonstrated by recent work in conversational interactions with robots during social \cite{kim2024understanding, wang2024ain} and collaborative tasks \cite{gkournelos2024llm, Wang2023-vg}. 
Mahadevan et al. \cite{mahadevan2024generative} present GenEM, a framework for generating control code with LLMs, utilizing a pre-defined robot skill library, such as ``move head'' and ``change light''.
GenEM is implemented on a mobile robot with a head-neck embodiment—primarily expressing through head movements, light changes, and directional motion, while \abbv operates on a humanoid platform capable of more sophisticated expressions, including human-like arm movements and individual finger articulations. This humanoid embodiment allows for richer and more nuanced behaviors, combining hand gestures and body language to enhance expressiveness. Furthermore, unlike GenEM, which focuses on arranging sequences of high-level pre-defined skills, \abbv leverages LLMs to directly produce complex hand and finger trajectories with minimal examples. Additionally, qualitative results from our user study offer insights into the impact of movement patterns, hand, and finger positioning on the human perception in terms of naturalness and understandability.\looseness=-1



\para{LLMs for robotic sequence optimization}
Recent studies have demonstrated that pretrained LLMs possess the capability to autoregressively generate complex numerical sequences applicable to robotics. Mirchandani et al.~\cite{mirchandani2023large} show these models' capabilities in tasks ranging from the extrapolation of simple motion sequences to the enhancement of reward-conditioned trajectories in elementary control tasks such as CartPole. Similarly, Wang et al.~\cite{Wang2023-vg} illustrates that LLMs could generate low-level control commands enabling quadrupedal robots to walk by improving joint trajectories with historical data. Extending these capabilities to vision-based tasks, Palo et al.~\cite{di2024keypoint} employs Vision Transformers to extract 3D keypoint tokens from images, and utilize expert demonstrations as contexts for few-shot imitation learning with LLMs.\looseness=-1

Distinct from prior works targeting at generating manipulation or locomotion policies, \abbv focus on the domain of human-robot interaction. In addition, \abbv is designed to generate novel sequences unlike approaches aimed at few-shot adaptation from the single-task expert demonstrations~\cite{Wang2023-vg, di2024keypoint}. Furthermore, we explore the potential use of natural language feedback from humans to directly refine these sequences, thereby broadening the practical applications of LLMs in robotics.

\section{EXPRESSIVE MOTION SEQUENCE GENERATION}
\label{sec:method}

\para{Overview} As illustrated in Figure~\ref{fig:overview}, \abbv takes user language instruction robot image observation $o_{\text{in}} \in O$ and/or $l_{\text{in}} \in L$ as input and outputs a continuous-valued motion sequence for the humanoid robot, in the form of $\tau = \{s_1, \ldots, s_{T}\}$. We use $T=10$ in this work. \abbv also incorporates human feedback. At iteration $i \in \{1, 2, \ldots, i_{\text{max}}\}$, human feedback $f^{(i)}$ can be provided to improve the current motion sequence $\tau^{(i)}$ and produce a new motion sequence $\tau^{(i+1)}$. The motion sequence is executed on the humanoid robot with inverse kinematics and trajectory interpolation and tracking. 

\begin{figure}
\centering
\includegraphics[width=0.75\linewidth]{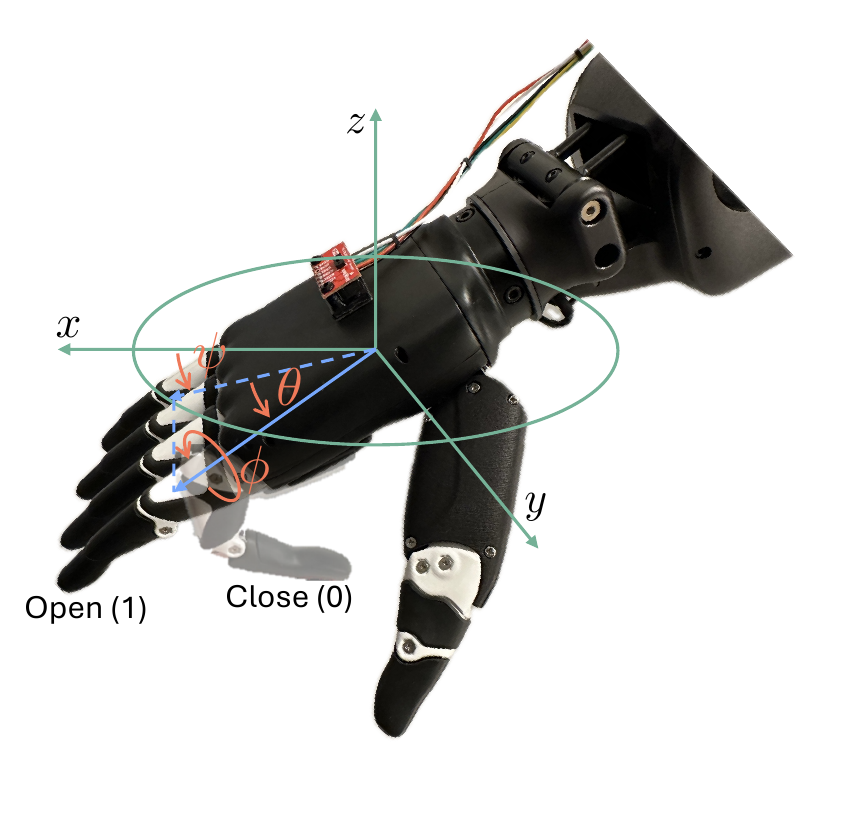}
 \vspace{-0.9cm}
\caption{Motion sequence representation.}
 \vspace{-0.5cm}
\label{fig:hand}
\end{figure}

\para{Motion Sequence Representation} The state at each timestep $s_t$ in $\tau$ is represented by 22 real values. We use six quantities to parameterize the motion sequence: left/right-hand Cartesian position (3D), orientation represented by Euler angles (3D), and opening/closing of each finger (5D), as shown in Figure~\ref{fig:hand}.

\para{Social Context Analysis} The inputs, including robot image observation $o_{\text{in}}$ and/or user language instruction $l_{\text{in}}$, provide important social context to which the robot needs to respond. For example, $o_{\text{in}}$ could be an image of a person solving a math problem on a whiteboard. The prompt is then to ask the LLM agent to describe what it sees and how to respond with only gestures. Input could also be as simple as a language instruction, such as ``Express confusion with only gestures". The inputs are fed to the first VLM agent $f_1(o_{\text{in}}, l_{\text{in}})$ with chain-of-thought (CoT) prompting to obtain the social context analysis $l_{\text{a}}$, and the corresponding gesture $\langle g \rangle$. Formally, $[l_{\text{a}}, \langle g \rangle] = f_1(o_{\text{in}}, l_{\text{in}})$. For example, $l_{\text{a}}$ could be ``I see a person excitingly solving a complicated math problem on a whiteboard. I would communicate approval and encouragement for the person’s achievement", and $\langle g \rangle$ could thereby be ``$\langle$thumbs-up$\rangle$".

\para{Human Demonstrations} We maintain a set of human demonstrations $\mathcal{D} = \{(\langle g_{\text{demo}} \rangle, l_{\text{demo}}, \tau_{\text{demo}})_{d}\}_{d=1}^{D}$, where $\langle g \rangle$ is the gesture name, $l_{\text{demo}} \in L$ is the one-line description of the gesture, and $\tau_{\text{demo}}$ is the motion sequence of the demonstrated gesture. Specifically, we find that with just two human demonstrations ``idle'' and ``right-hand wave'' ($D=2$), it provides enough information for the downstream generation tasks. We use Apple Vision Pro to collect human demonstrations and transform them into the same representation as the one that LLMs are required to generate. 

\para{LLMs as a Motion Sequence Generator using In-Context Learning} We denote the second LLM agent as $f_2$. $f_2$ takes in the gesture $\langle g \rangle$ together with instructions about the coordinate definition, motion sequence representation, as well as CoT prompts to facilitate the generation. Conditioned on the human demonstrations $\mathcal{D}$, it then outputs the initial motion sequence $\tau^{(1)}$ in the aforementioned representation. Formally, $\tau^{(1)} = f_2(\langle g \rangle ; \mathcal{D})$.

\para{Iterative Improvements from Human Feedback} \abbv incorporates human feedback (HF) in a natural language format. We denote the HF version as \abbvh. The third LLM agent takes in the history of the generated motion sequences $\tau$ and human feedback $l_f$ and outputs the new version of the motion sequences. Formally, $\tau^{(i+1)} = f_3(\tau^{(1)}, l_f^{(1)}, \tau^{(2)}, l_f^{(2)}, \ldots, \tau^{(i)}, l_f^{(i)})$, where $i \in \{1, 2, \ldots, i_{\text{max}}\}$. For example, this $l_f$ could either command an explicit modification in the current motion, such as ``put your hands higher'', or suggest a high-level improvement, such as ``add some random motion''.

The motion sequences for \abbv and \abbvh are generated by sampling OpenAI's GPT-4o (gpt-4o-2024-05-13) APIs for text completion~\cite{achiam2023gpt}. We include all the prompts in the supplementary material.

\section{USER STUDIES}

\begin{figure*}[h]
    \centering
    \includegraphics[width=0.99\linewidth]{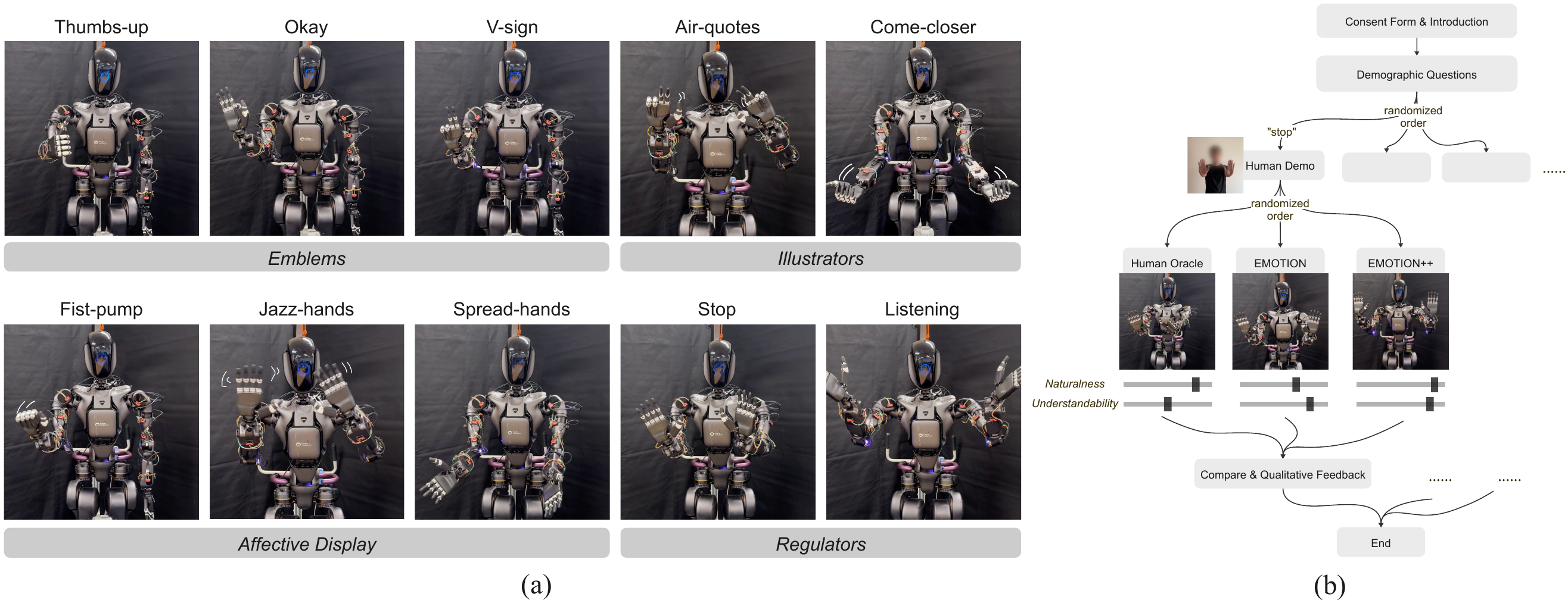}
    \caption{(a) 10 experimented robot expressive gestures under four non-verbal gesture categorizations, i.e., emblems, illustrators, affective displays, regulators. (b) Illustration of the survey workflow.}
    \label{fig:gestures}
    \vspace{-0.3cm}
\end{figure*}

\subsection{Research Questions and Hypotheses}
As describe in Section~\ref{sec:method}, \abbv has two main steps: social context analysis and motion sequence generation. Since the social context analysis capability has been explored in the existing work \cite{mahadevan2024generative} and to eliminate unnecessary randomness during the LLM inference time, we focus on investigating the motion sequence generation capability of LLMs and fix the set of gestures to be investigated as detailed in Section~\ref{sec:exp_mat}. To understand the effectiveness of the proposed method, we aim to study the following research questions: 
\begin{enumerate}
    \item \textbf{RQ1:} How well are the generative behaviors (\abbv) perceived when compared to the human oracle behaviors?
    \item \textbf{RQ1:} Whether and to what extent does adding the human feedback for \abbvh improve the perception of the gestures, compared to \abbv?
\end{enumerate}

Therefore, our hypotheses are:
\begin{enumerate}
    \item \textbf{H1:} the perception of the \abbv behaviors would not differ significantly from the oracle human behaviors.
    \item \textbf{H2:} the \abbvh behaviors would be perceived as more natural and understandable compared to the \abbv.
\end{enumerate}

\subsection{Experiment Materials}
\label{sec:exp_mat}
To generate behaviors for evaluation and comparison between models, we used the GR-1 humanoid robotic hardware from Fourier \cite{GR1}. Expressive gestures were either generated by models or recorded by a human operator, then verified and selected by a human researcher within the robot's simulation platform. For \abbvh, a human researcher provided feedback during this stage until the simulated behaviors aligned with the expected outcomes with a maximum feedback iteration $ i_{\text{max}}=5$. Finally, the gestures were deployed on the humanoid robot for recording, during which we minimized the difference in lighting and background. 

Following the established categorization of non-verbal communication~\cite{ekman1969repertoire}, we investigate four types of expressive gestures: (1) \textit{emblems}, non-verbal gestures that can generally be translated directly into words. (2) \textit{illustrators}, motions that complement verbal communication by describing, accenting, or reinforcing the speech. (3) \textit{affect displays}, motions that carry an emotional meaning or display affective states. (4) \textit{regulators}, non-verbal messages that accompany speech to control or regulate the verbal communication. We collect a total of 10 gestures covering the four categories for deployment and evaluation, including ``thumbs-up'', ``okay'', ``v-sign'' (emblems); ``air-quotes'', ``come-closer'' (illustrators); ``fist-pump'', ``jazz-hands'', ``spread-hands'' (affect displays); ``stop'', ``listening'' (regulators), as shown in figure \ref{fig:gestures}a. We include the video clips of all the gestures in the supplementary material.

\subsection{Survey Design}

To explore the research questions and compare differences in the perception of human oracle behaviors and generative behaviors (including \abbv and \abbvh), we conducted an online survey. In this survey, participants watched videos of the generated behaviors and evaluated how understandable and natural the gestures are perceived.

We employed a within-subject study design. Each participant viewed gestures across three conditions: (A) human oracle, (B) \abbv, and (C) \abbvh, presented in a randomized order. After each video, participants rated the perceived naturalness and understandability of the gesture they observed. Following this, participants were asked to directly compare the three behaviors and provide qualitative reasoning for their choices.
Participants view and evaluate 10 gestures mentioned in Figure~\ref{fig:gestures}a in a randomized order, to counterbalance the learning effect of the task. 
To help the participants understand the meaning of each gesture, we provide a video of a human performing the gesture along with a description for context. The process of the experiment is illustrated in figure \ref{fig:gestures}b.


\para{Measures} For each behavior, we included two quantitative metrics: \textbf{Naturalness} and \textbf{Understandability}. Participants self-reported their ratings on 7-point Likert scales, with ``7'' indicating ``completely natural'' and ``completely understandable''. Additionally, we collected optional qualitative feedback for each gesture, allowing participants to provide reasoning and insights behind their choices.

At the start of the study, we collected demographic data, including gender, age, ethnicity, general attitudes toward robots, and the general use of gestures for expression. These demographic features were included in the analysis to understand their potential effects on the results.

\subsection{Participants}
We recruited 30 participants using emails and announcements distributed within the organization\footnote{The study is exempt under the organization's Human Study Review Board criteria. This study fits under the research involving benign behavioral interventions and collection of information from adults with their agreement (CFR 46. 104 (d) (3))}. Responses were filtered based on the time taken to complete the task, excluding those that took less than eight minutes, as well as any incomplete responses. This process resulted in 22 valid participants ($N=22$). Among them, 6 are female, 15 are male, and one participant did not disclose their gender. The participants' ages range from 18 to 54 years. In terms of ethnicity, 18 participants self-identified as Asian, 2 self-identified as White or Caucasian, and 2 preferred not to disclose their ethnicity.

\section{RESULTS}

\begin{figure*}[ht]
    \centering
    \includegraphics[width=0.8\linewidth]{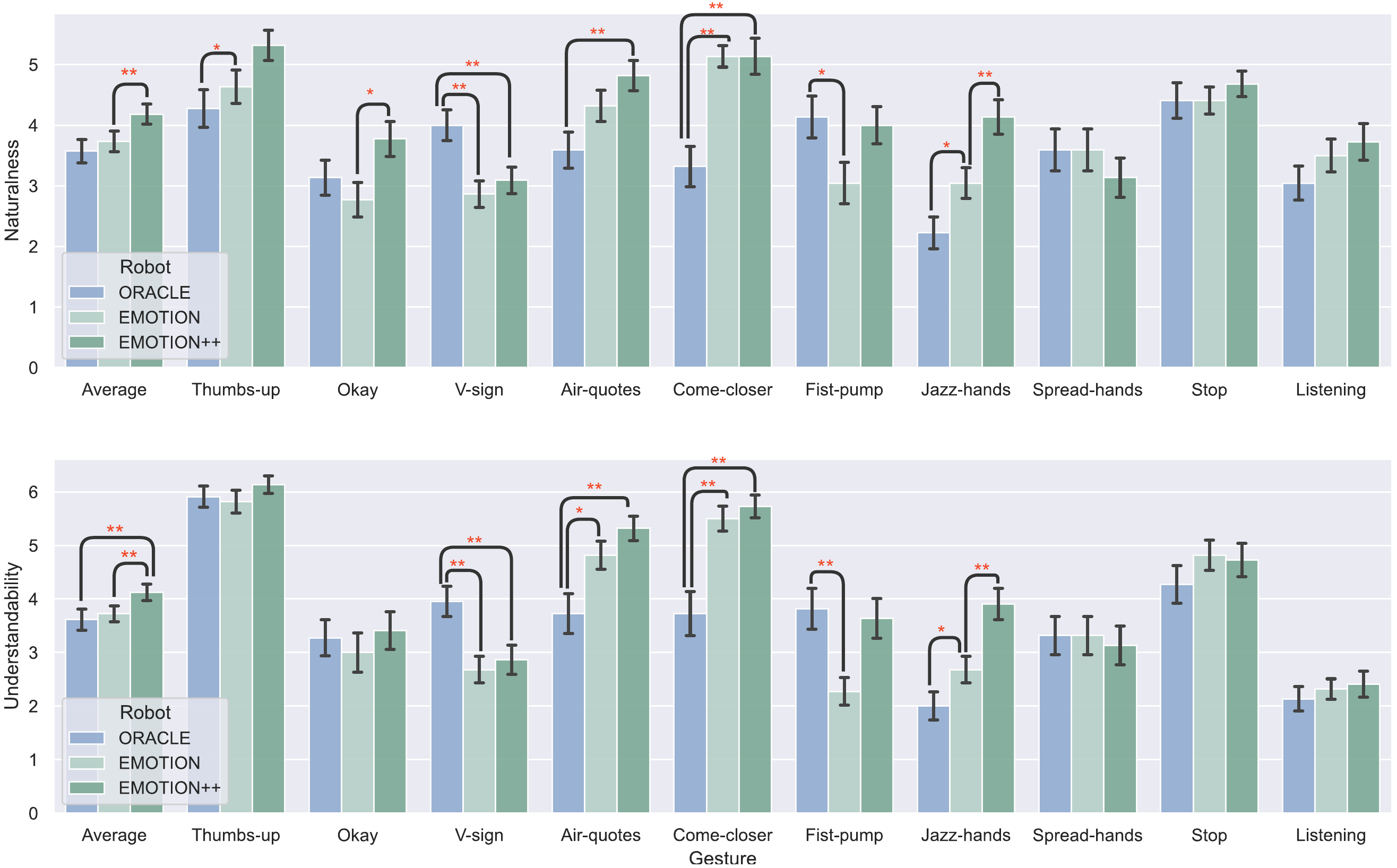}
    \caption{Users' rated scores of understandability and naturalness for generated robot expressive behaviors, separated by gestures. \textcolor{red}{*} and \textcolor{red}{**} indicate statistical significance with one-way ANOVA analysis (\textcolor{red}{*} indicates $p<0.05$ and \textcolor{red}{**} indicates $p<0.01$). The error bars indicate standard error (SE) of the mean.
    }
    \label{fig:bar}
\vspace{-0.4cm}
\end{figure*}


\subsection{Quantitative Results}

We perform quantitative analysis on users' self-reported ratings of naturalness and understandability for each generated behavior. Figure~\ref{fig:bar} plots the ratings of perceived naturalness (top) and understandability (bottom) for individual gestures, and for three different conditions: Human oracle (blue), \abbv generated behaviors (light green), and \abbvh generated behaviors (dark green).

To test H1, we run the one-way ANOVA test to compare each and combined gestures between perceived human oracle behaviors and \abbv generated behaviors. Combining all the gestures, \abbv behaviors does not statistically differentiate from human oracle behaviors (Naturalness: $p = 0.267$, Understandability: $p = 0.528$).
Thus, H1 is supported.

To test H2, we use one-way ANOVA analysis to compare the perceived naturalness and understandability between \abbv and \abbvh. 
Combining the gestures altogether, \abbvh is rated as significantly more natural and understandable than \abbv (naturalness: $p = 0.0014$, understandability: $p = 0.019$). 
Looking at the individual gesture, the significance of the differences between \abbvh and \abbv exist in the gestures ``fist pump'', ``jazz hand'', and ``okay'' (only naturalness). 
Thus, H2 is supported.

To further understand the quality of the generated motions integrating human feedback, we compare \abbvh to human oracle with one-way ANOVA tests. Combining all the gestures, \abbvh behaviors received significantly higher ratings for understandability than human oracle behaviors ($p = 0.003$). Perceived naturalness of \abbvh behaviors are also rated higher without statistical significance revealed. 
For individual gestures, \abbvh generated behaviors for ``air quotes'', ``come closer'', ``thumbs-up'' are perceived significantly more natural than human oracle ones, among which ``air quotes'' is also perceived as significantly more understandable. 
On the other hand, human oracle gesture ``v-sign'' received significantly higher ratings in both naturalness and understandability than \abbvh generated gestures. The results above suggested that the generated motions can not only match the performance of human oracle expressions, but may surpass them in many gestures, especially after incorporating human feedback.

\begin{figure*}[ht]
    \centering
    \includegraphics[width=1\linewidth]{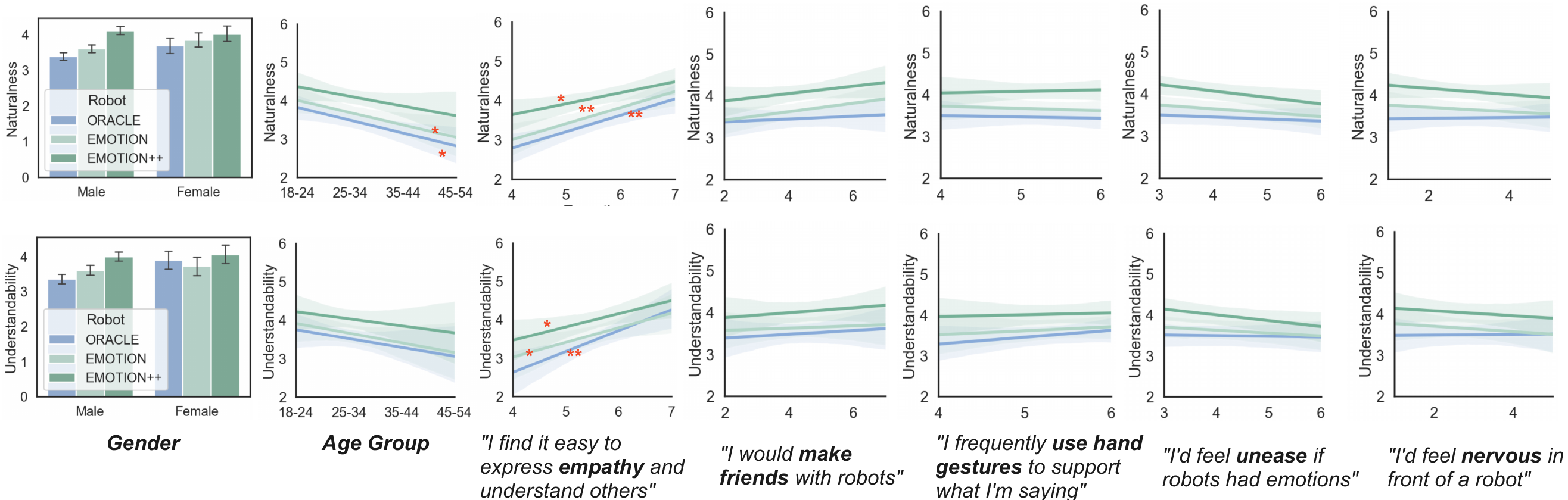}
    \caption{Correlations between the perceived naturalness / understandability and demographic variables, including participants' age, gender, general attitude toward robots,  empathy and frequency of using hand gestures in their daily life. (\textcolor{red}{*} indicates $p<0.05$ and \textcolor{red}{**} indicates $p<0.01$, suggesting significance in the correlation). The shaded areas indicate $95\%$ confidence interval, while the error bars indicate standard error (SE).
    }
    \label{fig:demographic}
    \vspace{-0.4cm}
\end{figure*}

To understand the effect of participants' backgrounds on their perceptions of robotic behaviors, we conducted a linear regression to examine the correlations between perception metrics (understandability, naturalness) and background variables, including age, gender, general empathy towards others, intent to make friends with robots, frequency of using hand gestures, unease with robotic emotions, and nervousness about interacting with robots in general, as shown in Figure~\ref{fig:demographic}.
We found that participants' self-rated empathy levels were significantly positively correlated with the perceived understandability and naturalness of the robot's generated expressions. The higher the empathy level, the more natural and understandable the gestures were perceived by the participants (Naturalness: Oracle: $p=0.0006$, \abbv: $p=0.0007$, \abbvh: $p=0.023$; Understandability: Oracle: $p=0.00027$, \abbv: $p=0.011$, \abbvh: $p=0.022$). This finding aligns with psychological literature, which suggests that high empathy levels are associated with enhanced narrative comprehension \cite{gallagher2017empathy}.
On the other hand, participants' age was negatively correlated with the perceived naturalness of the behaviors (Naturalness: Oracle: $p=0.033$, \abbv: $p=0.039$, \abbvh: $p=0.109$; Understandability: Oracle: $p=0.237$, \abbv: $p=0.193$, \abbvh: $p=0.344$).
We did not find any statistically significant correlations for the other variables.


\subsection{Qualitative Results}

To further reveal the insights of the perceived differences between the generative models and human oracle for individual gestures, we perform qualitative analysis of participants' reasoning comparing the three behaviors for each gesture.
To identify the cluster of ideas that affect participants' perception, we use thematic analysis~\cite{kiger2020thematic} to code participants' quotations, and identify clusters of codes to formulate themes and summarize key findings under each theme. We identify three main themes and twelve sub-themes, including robot variables, human perception, and contextual factors.



\subsubsection{Robot variables}
We identified 114 quotations discussing how robotic variables affected participants' perceptions. The quotations cluster under five sub-themes:

\paragraph{Position of hands}
For ``v-sign'' gesture, seven quotations expressed that a human oracle has a better hand position, where the robot raises the hand up and with the elbow angle resembling a human's.
For other gestures, there are similar comments on that the higher \textbf{hand position} is more perceivable and natural.
For ``listening'' gestures, 10 quotations mentioned hand position affecting the interpretation, with participants preferring the hands to be up close to the ears. Several said \abbvh has the best hand position in this term.
Incorrect hand positions could confuse the gesture interpretation. For example, for the ``listening'' gesture, participants commented that the positions of the hands are not close enough to the ears, and could be interpreted as gestures like ``surrender'' or ``stop''.

Besides position, the \textbf{orientation of the hands} also affect naturalness and engagement. For ``come closer'' gesture, P1 mentioned that the human oracle gesture ``is more toward me, it feels more engaging''.
For ``spread-hands'' gesture, participants prefer the hands to face upward. For gestures that engage two hands, the \textbf{configuration of the two hands} could affect how it is perceived, including the distance, whether they are symmetrical, and the synchronization of the motions. For example, P17 commented that the robot could ``widen the hands more to look more natural'' for ``spread-hands''.

\paragraph{Movement pattern}
24 quotations fit under the movement, with participants commenting on the jerkiness, the trajectory path, and direction and scale of the motions.
Most participants preferred ``taking the shortest and most direct path'' which is similar to human, among which three participants mentioned they preferred \abbvh in the ``okay'' gesture for this reason.

However, for some gestures, such as the ``fist pump'', ``jazz-hands'', and ``thumbs-up'', the participants preferred some \textbf{subtle movements}. P3 mentioned liking the subtle hand shaking at the end for the \abbvh gesture. For ``thumbs up'', P13 and P18 liked the subtle motions added by the human oracle at the end of the trajectory that ``seems to add emphasis'' to the expression. 
The \textbf{trajectory path} and the movement pattern affect perceived naturalness. For example, for the ``stopping'' gesture, both P10 and P14 commented that the Oracle has a clear push forward motion at the end of the gesture, which ``gives a more clear sign of stopping''. P18 mentioned that the movement of the human oracle for the ``fist pump'' appears more affirmative than others.

Several participants mentioned that the \textbf{jerkiness} of the motion affected the naturalness, especially for the ``thumbs-up'' gesture. Most participants mentioned that human oracle gestures are more jerky. The \textbf{timing} of the movement and the \textbf{coordination} of the motions also affected how realistic the gesture is perceived.
For the ``okay'' and ``v-sign'' gestures, participants commented that ``humans tend to not finish making the gesture [or finger pose] until the hand is in the ideal position.'' (P11)
P17 mentioned for the ``stopping'' gesture, when to open the hands affects naturalness. While humans only open the fist when the hands are in certain position, robots open the hands directly in the beginning.

\paragraph{Arm and shoulder}
13 quotations are discussed about \textbf{movement of the arm and shoulder} specifically, with 10 of them coming from the ``come closer'' gesture. 
While seven of the quotes liked the \abbvh making arm motions along with the fingers, which makes it more ``human-like'', and ``helps to convey to come closer''; three quotes found the arm motion was ``too much'', ``unnecessary'', and ``distracting''. 
For ``stop'' gesture, participants discussed that the angle of arm extension affects the interpretation, since the arm was not fully extended outward, thus difficult to interpret.


\begin{table*}[tb]
\centering
\caption{Computation time (in seconds) averaged over 10 independent runs. $\pm$ represents the standard deviation. The time shown includes the generation of reasoning text.}
\vspace{-4mm}
\begin{center}
\begin{adjustbox}{width=1.0\linewidth}
\begin{tabular}{l|ccccccccccc}
\toprule 

    
& Thumbs-up & Okay & V-sign 
& Air-quotes & Come-closer & Fist-pump 
& Jazz-hands & Spread-hands & Stop & Listening & \textbf{Average}\\
\midrule
Initial sequence       & 28.7 $\pm$ 4.0 & 33.1 $\pm$ 2.6 & 26.8 $\pm$ 3.3 & 24.0 $\pm$ 4.0 & 28.0 $\pm$ 5.2 & 26.6 $\pm$ 3.9 & 23.8 $\pm$ 2.5 & 24.9 $\pm$ 4.9 & 25.2 $\pm$ 3.4 & 26.6 $\pm$ 2.4 & 26.8\\
Single-round HF       & 24.4 $\pm$ 11.8 & 24.8 $\pm$ 2.0 & 19.9 $\pm$ 2.7 & 20.6 $\pm$ 3.4 & 20.3 $\pm$ 3.7 & 21.5 $\pm$ 8.1 & 18.4 $\pm$ 3.1 & 19.2 $\pm$ 3.2 & 20.2 $\pm$ 2.7 & 22.3 $\pm$ 3.7 & 21.2\\
\bottomrule
\end{tabular}
\end{adjustbox}
\end{center}
\label{table:latency}
\vspace{-7mm}
\end{table*}
    

\paragraph{Finger pose}
22 quotations mentioned the problems with finger pose especially with Human Oracle, where some \textbf{fingers fail to bend or make contact with each other}. For ``okay'' gesture, participants commented that it is essential to make two fingers contacting each other and forming the ``O'' shape in the hand for delivering the meaning, while the robot had hardware limitations thus struggling to do so. For gesture ``v-shape'', participants wished the fingers to be more apart to resemble the ``v'' shape. Failing to do so may confuse the gesture interpretation. For example, P22 mentioned ``the fingers feel like they are smoking cigarettes'' for the ``v-sign''.
For ``air quotes'' gesture, P11 commented that human would not fully fold the fingers but all the robots did, which made it unnatural.\looseness=-1 

\paragraph{Speed}
14 quotations mentioned that the speed of the gestures of all three robots is much slower than when a human is performing it, thus affecting the perceived naturalness. 8 quotations are specifically commenting on the slowness of the ``air quotes'' gesture.

\subsubsection{Human perception}
There are 14 quotations specifically commenting on \textbf{naturalness} of the gestures, while referring to them as ``natural'', ``human-like'', ``[not] awkward / strange''. 
30 quotations discussed the understandability of the gestures.
14 of them mentioned the \textbf{understandability} issue of the gestures where it could be interpreted as completely different gestures. For example, several commented that ``listening'' gesture could be confused with ``stop'', ``giving up'', and ``hugging'' (P10, P16).
``jazz-hands'' could be interpreted as ``saying hi'', or ``struggling'' (P1, P10).
``Stopping'' gesture felt like the robot wanted to touch or shove something (P22, P13).

Three quotes mentioned about the \textbf{engagement} of the gestures, where all of them thought Human Oracle gestures are more engaged due to the movement patterns.
Four quotes discussed the \textbf{emotion} of the expressions. For ``jazz-hands'' especially, participants mentioned they although they might not perceive the specific gestures, they could feel the positive emotions from the \abbvh gestures, and describe them as ``the robot looks happy'' (P22), and ``feel like dancing''(P1).

\subsubsection{Contextual factors}
Six quotations were found discussing the context-related factors that have affected participants' perception of the gestures.
Three of them thought the human demonstration in the instruction had some issues for the gesture ``listening'', ``spread hands'', and ``come closer''. P6 mentioned that the human demonstration of ``spread hands'' can be interpreted as ``I don't know'' more than ``apology'' or ``regret''. P6 also wished the human weren't holding the elbows fixed for the ``come closer'' gesture.
Two participants thought the gesture could have different interpretation under different \textbf{contexts}, and could depend on personal experience. 
One participant mentioned that ``jazz-hands'' is not a familiar gesture and they wouldn't know without seeing the prompt.


\section{Additional Experiments}
\label{sec:experiment}





\para{Computation Time} The computation time of \abbv mainly depends on two factors: model size and number of generated tokens. Since we are using OpenAI APIs to generate the response, it is also related to internet connection speed and available server resources at the moment. Therefore, the measured time could have significant variation. As shown in Table~\ref{table:latency}, \abbv achieves an average computation time of $26.8s$ for the initial motion sequence generation and $21.2s$ for a single-round sequence generation after human feedback under our lab conditions. The computation time is still considered too long for real-time conversation. This issue could be mitigated by distilling a smaller LLM specialized in motion sequence generation and running inference locally instead of through APIs, which is an active research topic but not the main focus of this paper.\looseness=-1

\para{Human Feedback} One of the findings of this work is that it is possible to use natural language feedback to directly refine the motion sequence. For \abbvh, the human provide $1.9$ iteration of feedback averaged over 10 gestures. $21\%$ of the feedbacks are high-level commands about the subtle motion, such as ``add some nuanced motion'' or ``make the hands movement a little bit more exaggerated''. The rest are about the position of the hands, such as ``make both hands lower''.

\para{Failure Mode} The main failure mode is the feasibility to solve for collision-free trajectories with inverse kinematics. Since LLMs are not explicitly aware of the workspace, the motion sequence generated can occasionally be infeasible to track. In this case, we either regenerate the sequence or provide feedback to steer the trajectory. We experimented with joint-space representation but found that it is difficult for LLMs to understand the demonstrations and generate in the joint space.



\section{Discussion and Conclusion}

In this paper, we present \abbv and \abbvh, an LLM-based generative framework (incorporating human feedback) for generating expressive gestures for a humanoid robot, particularly focusing on hand and arm trajectories. Through an online user study, the behaviors generated with human feedback (\abbvh) outperform the human-oracle behaviors in perceived naturalness and understandability and well as generated behaviors without human feedback (\abbv). Variation remains across different gestures, and further analysis provides insights into differences in preferences for various gestures and robotic parameters.

Qualitative results indicate that factors such as a humanoid robot's hand positioning, movement patterns, arm and shoulder trajectories, finger poses, and movement speed all contribute significantly to the understandability and naturalness of expressive gestures. Future researches could benefit from a focus on these variables when creating demonstrations and designing prompts for behavior generation, considering the following design implications derived from the findings:

\begin{itemize}
    \item The robot’s hand positioning, orientation, and the physical configuration between the hands can convey gesture meaning. Generative models might improve behavior by explicitly articulating hand positions and two-hand spatial relationship that reflect human poses.
    \item While users generally prefer direct, efficient movements, adding subtle motion can enhance naturalness and engagement, though preferences vary. Generative models must balance subtle motions with the primary gesture trajectory to optimize interaction, potentially adapting to user feedback for more personalized expressions.\looseness=-1
    \item Finger poses play a crucial role, especially in emblematic gestures where specific shapes convey meaning, such as an ``okay'' or ``V-sign''. Future models should pay close attention to finger positioning for expressive accuracy while also considering the robot's hardware constraints.
    \item The timing and coordination of movements impact the human-likeness of gestures. 
    Future models should not only focus on trajectory paths for hands and arms but also on temporal coordination to better mimic humans.
\end{itemize}

The quantitative analysis revealed a correlation between participants' age, self-reported empathy levels and their ratings of understandability and naturalness. 
Future research on generative expression may consider participants' background variables when evaluating the perception of generated behaviors.\looseness=-1


We acknowledge certain limitations in our experiment, which we plan to address in future studies.
First, hardware constraints may restrict the robot’s expressive capabilities. For instance, the robot’s fingers lack subtle movements, which may limit the replication of certain human gestures. 
This limitation may have influenced participants' perceptions, and results could vary with alternative hardware. Second, the computation time is still considered too long for real-time usage. This issue could be mitigated by distilling a smaller LLM specialized in motion sequence generation and running inference locally instead of through APIs.
Third, human intervention during the prompting, feedback provision, and validation stages may introduce potential biases. To mitigate this, we provide all prompts and human feedback used in this work for transparency. Future work might assess model variance and reduce biases introduced by human preferences.

\balance
\bibliographystyle{IEEEtran} 
\bibliography{citation}

\end{document}